\documentclass[10pt,twocolumn,letterpaper]{article}

\usepackage{iccv}
\usepackage{times}
\usepackage{epsfig}
\usepackage{graphicx}
\usepackage{amsmath}
\usepackage{amssymb}

\usepackage{booktabs}
\usepackage{float}
\usepackage{paralist}

\usepackage{array}
\newcommand{\PreserveBackslash}[1]{\let\temp=\\#1\let\\=\temp}
\newcolumntype{C}[1]{>{\PreserveBackslash\centering}p{#1}}

\usepackage[pagebackref=true,breaklinks=true,letterpaper=true,colorlinks,bookmarks=false]{hyperref}

\usepackage[capitalize]{cleveref}
\crefname{section}{Sec.}{Secs.}
\Crefname{section}{Section}{Sections}
\Crefname{table}{Table}{Tables}
\crefname{table}{Tab.}{Tabs.}

\iccvfinalcopy 

\def\iccvPaperID{12514} 

\ificcvfinal\fi

\usepackage{xspace}
\newcommand{\mbf}{\mathbf}

\newcommand{\AvgDist}{AvgDist\xspace}

\newcommand{\tdfov}{3DFoV\xspace}
\newcommand{\GP}{GP\xspace}
\newcommand{\VAP}{VAP\xspace}

\newcommand{\threedpoint}{\ensuremath{\mbf{P}}\xspace}
\newcommand{\twodpoint}{\ensuremath{\mbf{p}}\xspace}

\newcommand{\cam}{\ensuremath{\mbf{c}}\xspace}
\newcommand{\eye}{\ensuremath{\mbf{e}}\xspace}

\newcommand{\eyecoordinatesystem}{\ensuremath{C^{eye}}\xspace}

\newcommand{\predictedgaze}{\ensuremath{\mbf{g_{p}}}\xspace}
\newcommand{\gtgaze}{\ensuremath{\mbf{g_{gt}}}\xspace}

\newcommand{\heatmap}{\ensuremath{\mbf{A}}\xspace}
\newcommand{\pseudoheatmap}{\ensuremath{\mbf{V}}\xspace}
\newcommand{\inoutindicator}{\ensuremath{\mbf{o}}\xspace}

\newcommand{\pointcloud}{\ensuremath{\mbf{P}}\xspace}
\newcommand{\cosinesimilarity}{\ensuremath{\mbf{c}}\xspace}

\newcommand{\headcrop}{\ensuremath{\mbf{I}_{h}}\xspace}
\newcommand{\headmask}{\ensuremath{\mbf{I}_{m}}\xspace}

\newcommand{\inputimage}{\ensuremath{\mbf{I}}\xspace}

\newcommand{\featuremap}{\ensuremath{\mbf{F}}\xspace}

\newcommand{\gazeembedding}{\ensuremath{\mbf{e_g}}\xspace}
\newcommand{\sceneembedding}{\ensuremath{\mbf{e_s}}\xspace}

\newcommand{\gazesubnetwork}{\ensuremath{\mathcal{G}}\xspace}
\newcommand{\featuresubnetwork}{\ensuremath{\mathcal{F}}\xspace}
\newcommand{\compressionnetwork}{\ensuremath{\mathcal{C}}\xspace}
\newcommand{\regressionnetwork}{\ensuremath{\mathcal{R}}\xspace}
\newcommand{\inoutnetwork}{\ensuremath{\mathcal{O}}\xspace}

\newcommand{\loss}{\mathcal{L}}
\newcommand{\losscoeff}{\lambda}

\newcommand{\mypartitle}[2][2.3]{\vspace*{-#1 ex}~\\{\noindent {\bf #2}}}
\newcommand\blfootnote[1]{%
  \begingroup
  \renewcommand\thefootnote{}\footnote{#1}%
  \addtocounter{footnote}{-1}%
  \endgroup
}

\begin{document}

\title{ChildPlay: A New Benchmark for Understanding Children's Gaze Behaviour\vspace*{-3mm}}

\author{Samy Tafasca*, Anshul Gupta*, Jean-Marc Odobez\\
Idiap Research Institute, Martigny, Switzerland\\
Ecole Polytechnique Fédérale de Lausanne, Switzerland\\
{\tt\small \{agupta, stafasca, odobez\}@idiap.ch}
}

\maketitle
\ificcvfinal\thispagestyle{empty}\fi

\blfootnote{* equal contribution}

\begin{abstract}
  \vspace*{-3mm}
   Gaze behaviors such as eye-contact or shared attention 
   are important markers for diagnosing developmental disorders in children. While previous studies have looked at some of these elements, the analysis is usually performed on private datasets and is restricted to lab settings. Furthermore, all publicly available gaze target prediction benchmarks mostly contain instances of adults, which makes models trained on them less applicable to scenarios with young children. In this paper, we propose the first study for predicting the gaze target of children and interacting adults.
   %
   To this end, we introduce the ChildPlay dataset: a curated collection of short video clips featuring children playing and interacting with adults in uncontrolled environments (e.g. kindergarten, therapy centers, preschools etc.), which we annotate with rich gaze information.
   %
   %
   %
   We further propose a new model for gaze target prediction that is geometrically grounded by explicitly identifying the scene parts in the 3D field of view (3DFoV) of the person, leveraging recent geometry preserving depth inference methods.  
   %
   Our model achieves state of the art results on benchmark datasets and ChildPlay. 
   Furthermore, results show that looking at faces prediction performance on children is much worse than on adults,  
   and can be significantly improved by fine-tuning models 
   using child gaze annotations.
   Our dataset and models will be made publicly available.
\end{abstract}

\vspace*{-5mm}

\section{Introduction}
\label{sec:intro}

    \vspace*{-2mm}

Gaze is a non-verbal cue that provides rich information about people.
%
In particular, it plays fundamental roles in social interactions and human communication,
like initiating interaction, showing attention, monitoring the floor,
and as such finds many applications in human interaction analysis
%
including  psychological studies and medical diagnosis.
%

In this regard, acquiring appropriate social gaze behavior skills
plays an important role in the cognitive developement of children.
Failure to learn such skills has been shown to be correlated with several
neuro-developmental disorders like  Autism Spectrum Disorder (ASD)~\cite{senju2009atypical}\cite{mundy1990longitudinal}.
This has led to the development of specific gold standard markers for the screening of these impairments,
e.g. by measuring deficits in the initiation of joint attention or shared attention~\cite{edition2013_dsm5}.

\begin{figure}[t]
    \vspace*{-2mm}
    \centering
    \includegraphics[width=\linewidth]{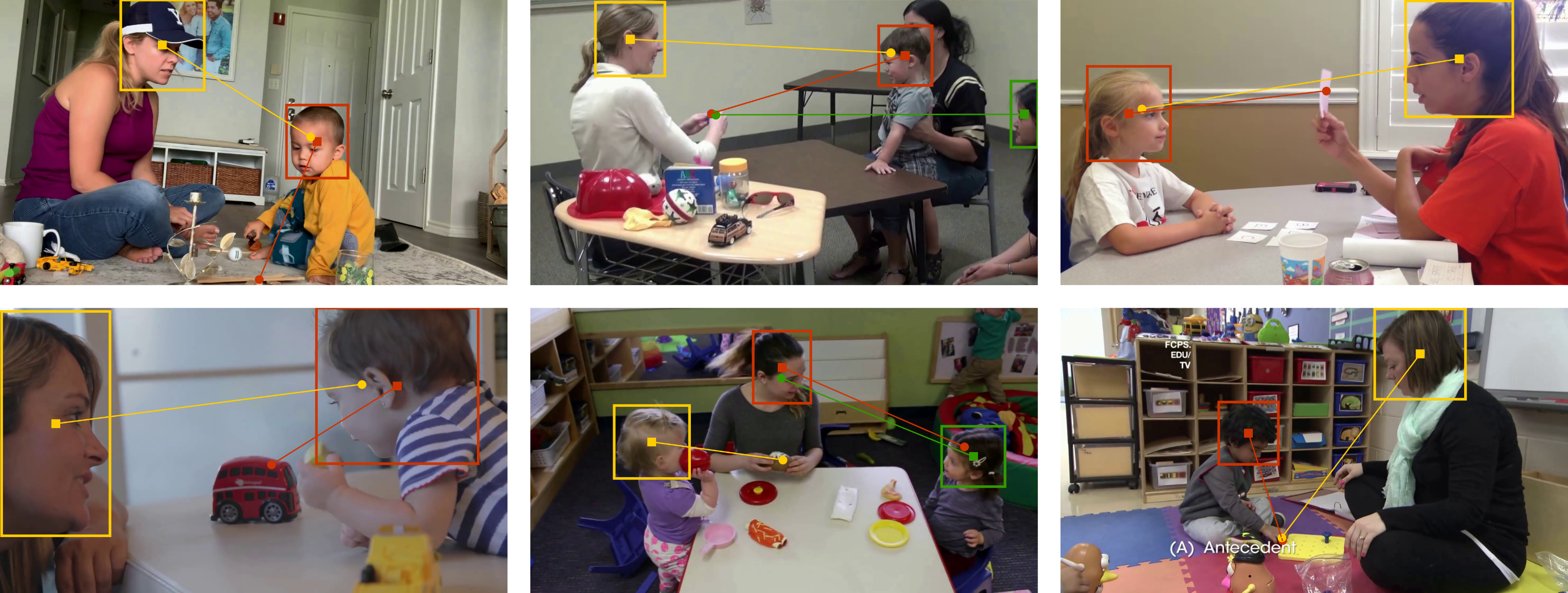}
    \vspace*{-5mm}
    \caption{Sample images from the ChildPlay dataset with head bounding box and gaze point annotations. Such scenes strongly depart from existing gaze benchmarks (\eg standing adults). }
    \vspace*{-2mm}
    \label{fig:childplay-sample}
\end{figure}

\begin{figure}[t]
    \centering
    \includegraphics[width=\linewidth]{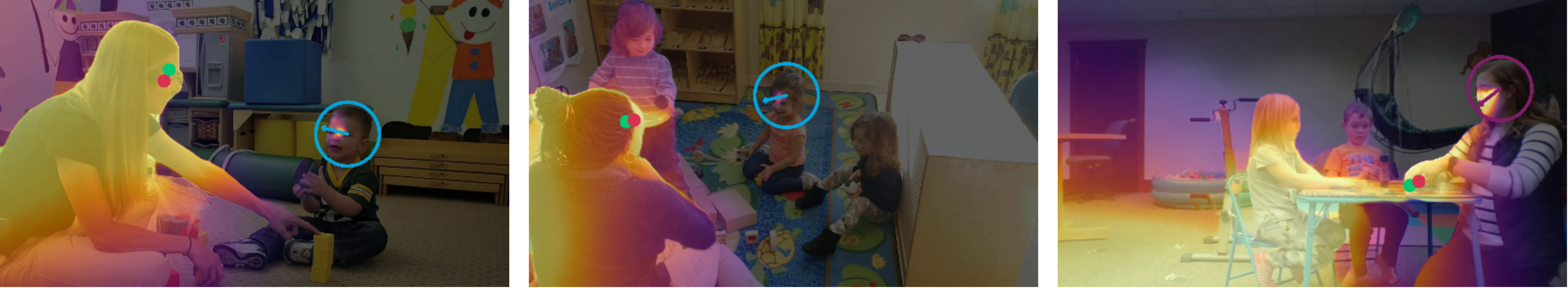}
    \vspace*{-5mm}
    \caption{Qualitative results of our geometrically grounded model on ChildPlay. Our 3D Field of View (3DFoV) highlights potential gaze targets, excluding objects where the depth does not match. 
    Gaze target predictions are given in green and GT ones in red.}
    \label{fig:qualitative}
    \vspace{-1em}
\end{figure}

\mypartitle{ChildPlay dataset.}
Due to this importance, several methods have been proposed to analyze children's gaze, especially for ASD
\cite{Rehg:CVPR:2013,Chong:IMWU:2017,Hashemi:TAC:2018,Rudovic2018,Anzalone:PRL:2019, shic2011limited, syeda2017visual, chong2017detecting, li2016modified}.
However, they are all tested on private datasets \cite{de2020computer} due to the sensitive nature of the data, 
hindering proper comparison across algorithms. 
Some of them may be accessible in anonymized formats (e.g. pose skeletons), but this makes them difficult to use for the study of gaze behavior.
While other datasets have gaze related labels 
involving autistic 
children \cite{billing2020dream, rehg2013decoding}, they are usually recorded in a fixed lab setting and 
with coarse annotations.
Alternatively, we could use standard public benchmarks like \cite{chong2020dvisualtargetattention} for learning gaze prediction models, but children and the physical situations and task performed 
(seating on the floor, playing with objects) 
would be severely underrepresented. 
It has been shown that training models 
on datasets with mainly adults can lead to a significant drop in performance when tested on children, e.g. for body landmark prediction
\cite{sciortino2017estimation}.
%
Given the importance of pose information in gaze prediction \cite{gupta2022modular, belkada2021pedestrians} and the difference in gaze behavior between adults and children \cite{franchak2016free}, there is a need for gaze annotated datasets featuring lower age groups in general settings.

To address this problem,  we introduce the ChildPlay dataset: a set of videos featuring children in free-play environments interacting with their surrounding. The dataset is rich in unprompted social behaviors, communicative gestures and interactions
and features high quality dense gaze annotations, including a gaze class to account for special scenarios that arise in 2D gaze following.
Further, the 2D gaze information can be used to model other attention-related behaviors like shared attention,
gaze shifts,  eye contact and fixation points with minimal processing.
To the best of our knowledge, we are the first to establish a more representative gaze dataset aiming to cover children. 
%
%

%
\mypartitle{3D Fiel of View (3DFoV) for gaze target prediction.}
Recasens et al.~\cite{recasens2017following} introduced this task (also called gaze following) for general scenes which aims to predict the image 2D gaze location of a person in the image. Since then, many works have embraced  this paradigm \cite{lian2018believe, chong2018connecting,zhao2020learning,guan2020_pose,nan2021predicting,jin2021multi}, 
proposing  new models exploiting temporal information \cite{chong2020dvisualtargetattention},
or exploiting further cues like depth \cite{Fang_2021_CVPR_DAM, bao2022escnet, gupta2022modular}.
Indeed, inferring and understanding depth is crucial, as it provides  information about the scene structure
enabling  geometric reasoning
and ruling out salient objects or people 
which fall along the 2D line of sight of 
a person, but are actually not visible to the person in the 3D space.
%
%
%
%
In this context, as most datasets do not have depth information,
some methods opted for pre-trained monocular depth (or disparity) estimators to extract scene depth cues~\cite{Fang_2021_CVPR_DAM, bao2022escnet, gupta2022modular}.
%
%
However, such algorithms~\cite{ranftl2019midas, yin2020diversedepth} typically estimate the
depth up to an unknown shift and scale factors
which often result in stretched and distorted scenes 
unsuitable for proper 3D analysis.
%

In this paper we provide a more geometrically grounded approach 
leveraging a new algorithm~\cite{patakin2022depth} addressing these points,
correcting shifts, and yielding geometry-preserving depth maps that can be used 
to derive a proper scene point cloud, and explicitly match the predicted 3D gaze vector with this point cloud to derive the 3DFoV of the person.
In experiments, we show that this 
method generalizes well, providing better cross-dataset performance. 

%
%
%
%
%
%

\mypartitle{New gaze metric: looking at heads precision (P. Head).}
Standard performance metric for gaze following either have no physical interpretation (the Area Under Curve, AUC), 
or may not provide rich enough information about performance, as is the case of 2D distance metrics (how far is a 2D gaze prediction from the GT).
%
In practice, one is more interested at semantics, e.g. how accurate is a model at predicting the category (person, body part, object) of 
image regions being looked at. 
As objects might be hard to annotate at scale,
in this paper, 
leveraging highly accurate head detectors and since heads are one of the most important gaze category in many applications (ex. child looking at clinician for ASD diagnosis~\cite{edition2013_dsm5}), 
we propose to exploit looking at head precision (P.Head) metric for performance evaluation. 
We show that this measure greatly varies across datasets and can have rather low performance, 
that the distance and P. Head metrics may disagree 
and performance on children is rather different than on adults. On ChildPlay, while children exhibit a better distance performance, their P. Head metric is much worse. 

\mypartitle{Contributions.}
Our main contributions are:
\begin{compactitem}
\item We introduce the ChildPlay dataset, a curated collection of clips recording children
  playing and interacting with adults in uncontrolled environments, annotated with rich gaze information;
  %
  %
\item 
  We propose a new model for gaze target prediction that relies
  on the explicit modeling of the 3DFoV by exploiting geometrically
  consistent inferred depth maps.
  %
  %
  \item We propose to use the Looking At Head Precision metric to characterize performance.
\end{compactitem}
Extensive experiments on the GazeFollow, VideoAttentionTarget and ChildPlay datasets demonstrate that our approach produces the best or state-of-the-art results, motivating further studies on the topic. 
The dataset and models will be made publicly available.


\section{Related Work}
\label{sec:soa}
\vspace*{-2mm}

We discuss works related to gaze target prediction and highlight the methods that use depth information.
We discuss datasets on gaze and children in Section~\ref{sec:gaze-datasets}.

\mypartitle{Gaze Target Prediction.}
Traditional methods for gaze following rely on a 2-branch architecture consisting of a scene branch to identify salient regions in the image and a human-centric branch to infer the general gaze direction of the target person \cite{nips15_recasens, chong2020dvisualtargetattention, lian2018believe, jin2021multi, Fang_2021_CVPR_DAM, gupta2022modular}. Various ideas have since been proposed in the literature to boost this typical architecture, namely, inferring a 2D gaze cone \cite{lian2018believe, gupta2022modular}, using multimodal information \cite{Fang_2021_CVPR_DAM, gupta2022modular, guan2020enhanced}, leveraging the temporal context \cite{chong2020dvisualtargetattention}, or improving computational efficiency for multi-person scenarios \cite{tu2022end, jin2021multi}. There are also other related tasks that incorporate semantic information such as detecting eye-contact \cite{marin2019laeo, belkada2021pedestrians}, inferring the gaze target object \cite{tomas2021goo, wang2022gatector}, or shared attention behavior \cite{fan2018inferring_videocoatt} to cite a few. 

\mypartitle{Gaze Target Prediction using Depth.}
Fang et al.~\cite{Fang_2021_CVPR_DAM} used a pre-trained monocular depth estimator to extract the scene depth.
They split this  map into three depth-based saliency maps depending  on the depth of the target person,
and used a pre-trained gaze estimation model to select the corresponding one for a coarse matching. 
Jin et al.~\cite{jin2022depth} attach auxiliary branches during training to predict scene disparity and predict a 3D orientation vector. However, these are not used as input to the model and depth is implicitly encoded in the features.
Hu et al~\cite{hu2022we} follow a similar strategy as ours but match a coarse predicted 3D gaze
vector with the derived scene point cloud.
Further, they mainly target the use of RGB-D images,
and their derived point cloud when dealing with RGB images is not geometrically consistent due to their pre-trained depth estimator~\cite{yin2020diversedepth}. 
Bao et al.~\cite{bao2022escnet} also derive a scene point cloud and attempt to correct it by using humans as reference objects.
However, they do not predict a 3D gaze vector and hence do not perform any explicit matching of predicted gaze and depth.
%
Gupta et al~\cite{gupta2022modular} treat the depth map as an input 
and hence  do not perform any matching of 3D gaze and depth.

\section{ChildPlay Dataset}
\label{sec:childplay}
%
In the following, we describe several aspects of the ChildPlay dataset i.e. data collection strategy, annotation protocol, statistics and comparison to benchmarks.

%


\subsection{Data collection strategy}
\label{sec:datacollection}

\vspace{-1mm}

\mypartitle{Data selection.}
We relied on the YouTube video search engine with queries like "children playing toys", "childcare center",
or "kids observation" to retrieve videos matching our aim \footnote{Around 10\% of our data have the CC BY license,
whereas the other have no license, i.e. they follow the default YouTube license.}.
To foster quality we only looked for videos with an aspect ratio of 16:9  and a resolution of 720p or 1080p.
We downloaded the audio files as well, although in many cases they are not produced by the scene (e.g. commentary).

\mypartitle{Clip selection.}
The scene context of our  videos ranged from childcare facilities and schools to homes and therapy centers.
%
As full videos contain many irrelevant parts,
we selected clips
with children and featuring interesting gaze movements and social interactions.
%
We also made sure that clips contain no scene cuts, blurriness,
large overlaid graphical items, heavy zooming or camera movement.
Finally, to foster diversity,
we limited the duration and number of clips taken from each video or Youtube channel. 
\mypartitle{Content.}
We obtain a dataset of 401 clips, mainly restricted to indoor environments, showing at least 1 child, but oftentimes include 1 or 2 adults and multiple children.
%
The age group of the children varies from toddlers to pre-teenagers. 
The dominant activity of children is "playing with toys", but the dataset also includes a few clips
containing other interactions such as behavioral therapy exercises.
%


\subsection{Annotation protocol}
\label{sec:annotationprotocol}

\vspace{-1mm}

\begin{figure}[t]
    \centering
    \includegraphics[width=0.9\linewidth]{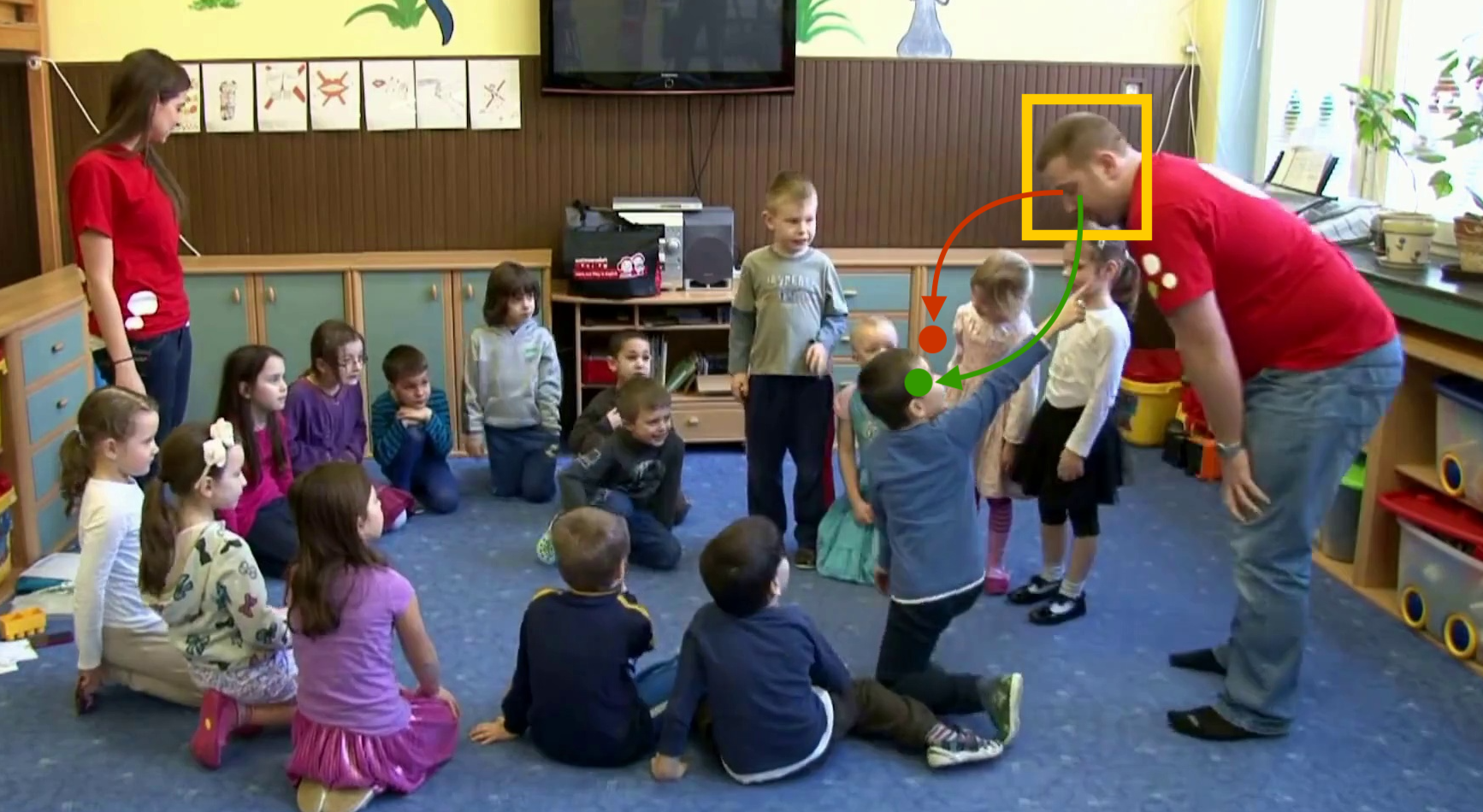}
    \caption{Example of two annotations close in $L_2$ distance but very different semantically (i.e. on different objects) and distant depth-wise. Green is correct, red is incorrect.}
    \label{fig:semantic-consistency}
    \vspace{-1em}
\end{figure}

We performed a dense annotation of gaze information.
In every clip we selected up to 3 people
%
and for each of them,  in every frame we annotated 
the head bounding box, a 2D gaze point, and a gaze label. We also provide the class label for adult vs child.
Two main points were taken into account to ensure high quality and confidence in the 2D annotations:
 enforcing semantically consistent 2D gaze annotations,
i.e. the annotated 2D location has to be on the object being looked at (cf Figure~\ref{fig:semantic-consistency}) in anticipation of a transition to semantically aware gaze evaluation metrics 
(as motivated by the LHP metric), 
and the introduction of a gaze label. 

The gaze label
addresses an important limitation with existing datasets, in which 
%
annotating a 2D gaze point is enforced in every frame,
with only a standard inside vs outside label to denote if the person looks within the frame or not. 
However, there are many situations where annotating 2D gaze points is highly challenging, if not impossible.
For example, when a person shifts attention from one location to another,
in the VideoAttentionTarget dataset we often observed that  intermediate frames during the shift
were annotated using the outside class (i.e. when the person slightly closes their eyes during the head movement)
which is inconsistent with the definition.
%
To avoid this, our gaze label was defined to include 7 non-overlapping classes to properly account for special scenarios: inside-frame, outside-frame, gaze-shift, occluded, eyes-closed, uncertain, not-annotated (precise definition in appendix).
%
%
%
In practice, inside-frame (85.3\%) and outside-frame (5.4\%) are the dominant classes,
but all other ones where a confident annotation can not be made still represent 9.3\% of the frames.

Finally, to evaluate the inter-annotator agreement in terms 
of 2D target localization we followed the usual practice.
We had 2000 instances being double coded, and evaluated the performance of a human (as prediction) against the other one (used as GT). Evaluation is reported in Table~\ref{tab:results-childplay} (under the "Human" baseline), demonstrating similar agreement as on other datasets (see \ref{tab:results-gazefollow-videoatt}).


%
%
%
%

\subsection{Statistics}
\label{sec:childplay-statistics}

\vspace{-1mm}

Table~\ref{tab:gaze-datasets} provides a summary of the main statistics  of our ChildPlay dataset.
%
It contains $401$ short clips averaging 10s, extracted from $95$ videos originating from $44$ different YouTube channels.
In total, there are around $258k$ annotation instances (i.e. $62\%$ for children) distributed across $120,549$ frames and annotated by 7 people using the LabelBox platform \cite{labelboxwebsite}.
Figure \ref{fig:childplay-sample} shows sample images along with the corresponding annotations.

\begin{figure*}
    \vspace*{-2mm}

    \centering
    \includegraphics[width=0.9\linewidth]{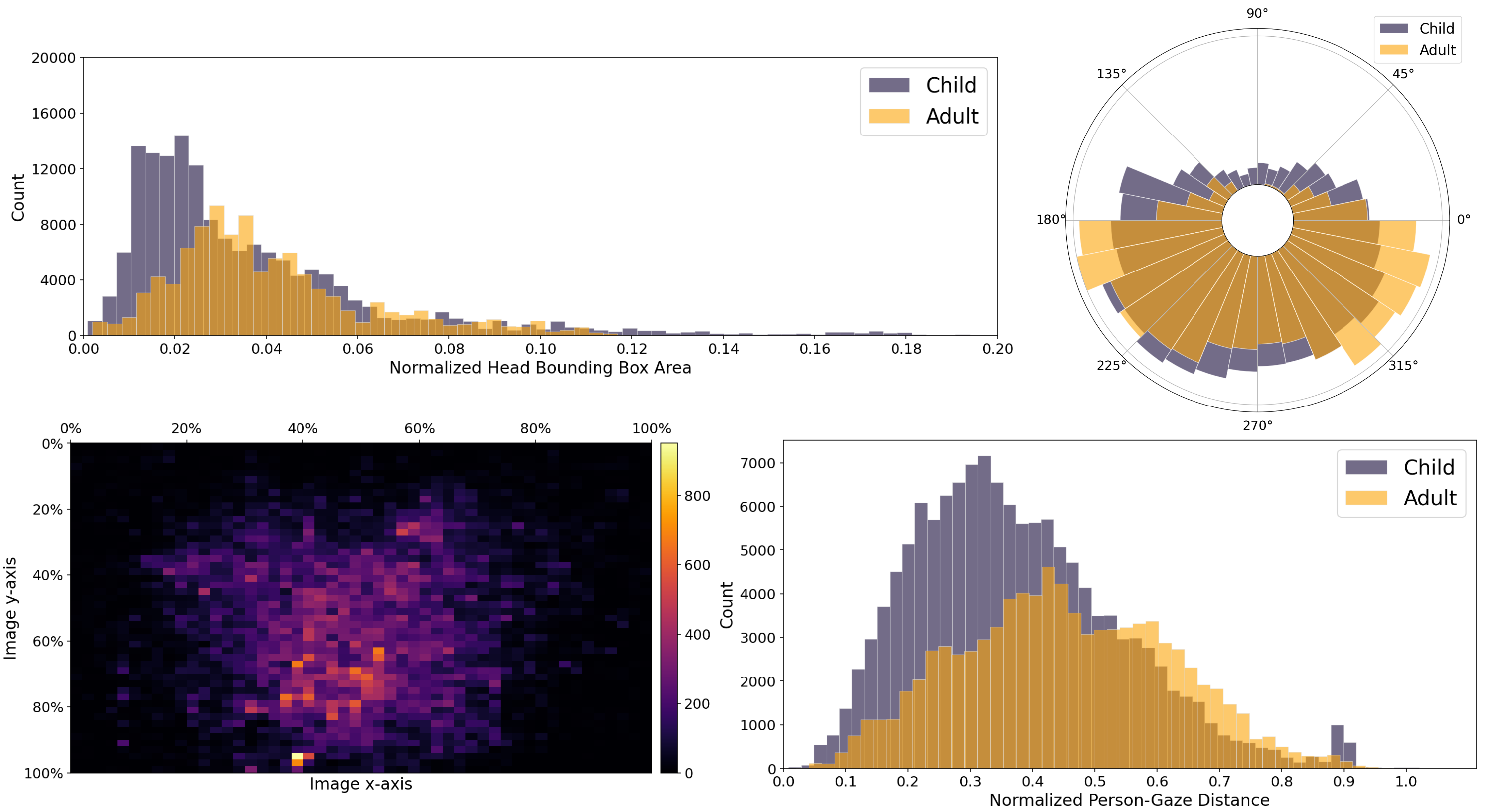}

    \vspace*{-1mm}
    
    \caption{Geometric statistics of the ChildPlay annotations.
    [Top-Left] Distribution of head bounding box area normalized w.r.t image size.
    [Top-Right] Distribution of gaze angles in the image frame.
    [Bottom-Left] 2D histogram of gaze points.
    [Bottom-Right] Distribution of the distance between each person (i.e. center of the head) and their gaze point, normalized w.r.t the image sides.
    }
    \vspace*{-2mm}
    
    \label{fig:childplay-stats}
\end{figure*}

Figure \ref{fig:childplay-stats} summarizes the distribution of various geometric quantities, highlighting major differences between children and adults. 
The distribution of head sizes covers a fairly wide range, and reflecting that people are located at various distances 
from the camera.
Moreover, we can notice that adults mostly look down to observe children and their activities, 
whereas children mostly look down at their toys with few instances where they raise their heads to gaze back at the adults, 
an important behavior difference which is also corroborated 
by the high difference in probability of looking at faces (see Tab.~\ref{tab:gt-looking-head}).
%
We can also observe that the distance from the head to the gaze point is typically between $10\%$ and $60\%$ of image side, 
with again children more focused on close targets than adults. 


\begin{table*}[t]
\footnotesize
\centering
{\footnotesize
\begin{tabular}{
p{0.17\linewidth}p{0.03\linewidth}p{0.06\linewidth}p{0.045\linewidth}p{0.05\linewidth}p{0.15\linewidth}p{0.275\linewidth}
}
    \toprule
    \textbf{Name} & \textbf{Type} & \textbf{Shows} & \textbf{Frames} & \textbf{Instances} & \textbf{Origin} & \textbf{Annotations} \\
    \addlinespace[2pt]
    \midrule
    
    GazeFollow \cite{nips15_recasens}  & image & - & $122,143$ & $130,339$ & SUN, MS-COCO, ImageNet, ... &  eye location $\cdot$ 2D gaze point $\cdot$ inside/outside \\
    
    \midrule
    
    VideoAttentionTarget \cite{chong2020dvisualtargetattention} & video & $50$ ($606$) & $71,666$ & $164,541$ & TV shows &  gaze point $\cdot$ inside/outside \\
    
    \midrule
    
    
    
    VideoCoAtt \cite{fan2018inferring_videocoatt} & video & $20$ ($400$) & $493,242$ & $138,203$ & TV shows &  shared gaze object bbox  \\ 
    
    \midrule
    
    DL Gaze \cite{lian2018believe} & video & $4$ ($86$) & $95,000$ & $6,348$ & Manual collection &  gaze point  \\ 
    
    \midrule
    
    UCO-LAEO \cite{marin2019laeo} & video & $4$ ($129$) & $18,000$ & $36,358$ & TV shows &  LAEO class  \\ 
    
    \midrule
    
    AVA-LAEO \cite{marin2019laeo} & video & $298$ & $1.4M$ & $172,330$ & Movies & LAEO class  \\
    
    \midrule
    
    GOO \cite{tomas2021goo} & image & - & $201,552$ & $172,330$ & Manual collection \newline Synthetic &  gaze point $\cdot$ gaze object $\cdot$ object bbox  \\
    
    \midrule
    
    ChildPlay & video & 95 (401) & $120,549$ & $257,928$ & YouTube &  gaze point $\cdot$ gaze class  \\  
    \bottomrule
\end{tabular}
}
\caption{
Summary of gaze estimation datasets. All datasets provide head bounding boxes (or pairs of them for LAEO).
}
\label{tab:gaze-datasets}
\vspace*{-2mm}
\end{table*}

\subsection{Comparison to other datasets}
\label{sec:datasetcomparison}

\vspace{-1mm}

\mypartitle{Gaze Datasets.}
\label{sec:gaze-datasets}
Table \ref{tab:gaze-datasets} presents an overview of existing gaze prediction datasets that are publicly available.
The two most related datasets are GazeFollow and VideoAttentionTarget.
GazeFollow \cite{nips15_recasens} is a large-scale image dataset featuring $130K$ independent instances, but it suffers from low resolution,
average annotation quality and lack of temporal context. Nevertheless, given its rich diversity, it remains a good dataset to use for pre-training.
VideoAttentionTarget \cite{chong2020dvisualtargetattention} is a recent video dataset
built from high resolution clips taken from popular TV shows.
Since it was extracted from 50 shows, the diversity of the scenes remains limited.
Furthermore, it contains mostly adults and has a very strong bias towards looking at people's faces,
as shown in Table~\ref{tab:gt-looking-head}.


%
%
Our ChildPlay dataset is far more balanced.
%
while also having 50\% more frames and twice as much scene variety.


Other datasets differ significantly from ChildPlay in their scope (beyond having very little children).
Several of them address attention related tasks (Co-attention  \cite{fan2018inferring_videocoatt},
looking-at-each-other (LAEO)  \cite{marin2019laeo}), specific settings like retail 
\cite{tomas2021goo},
or are much smaller in size and diversity hance can mainly 
be used for evaluation, not training \cite{lian2018believe}.

\mypartitle{Children Datasets.}
\label{sec:child-datasets}
The Multimodal Dyadic Behavior (MMDB) dataset \cite{rehg2013decoding}, the Self-Stimulatory Behaviors dataset (SSBD) \cite{rajagopalan2013self}, DREAM \cite{billing2020dream} and 3D-AD \cite{rihawi20173d} are all datasets meant to tackle different aspects of autism, be it stimming behaviors (arm flapping, head banging),
speech and vocalizations, communicative gestures (e.g. pointing, reaching, etc.) or gaze patterns (e.g. shared attention, eye-contact).
However, they are either anonymized, limited in terms of behaviors or restricted to lab environments (e.g. screening or therapy sessions). 
In contrast, ChildPlay boasts a higher diversity of scenes, people, gestures, viewing angles and lighting conditions. 
More information about children datasets in appendix.

\section{Model Architecture}
\label{sec:architecture}
\vspace*{-1.5mm}

\subsection{Approach overview}

\vspace*{-2mm}


\begin{figure*}
    \centering
    \includegraphics[width=0.95\textwidth]{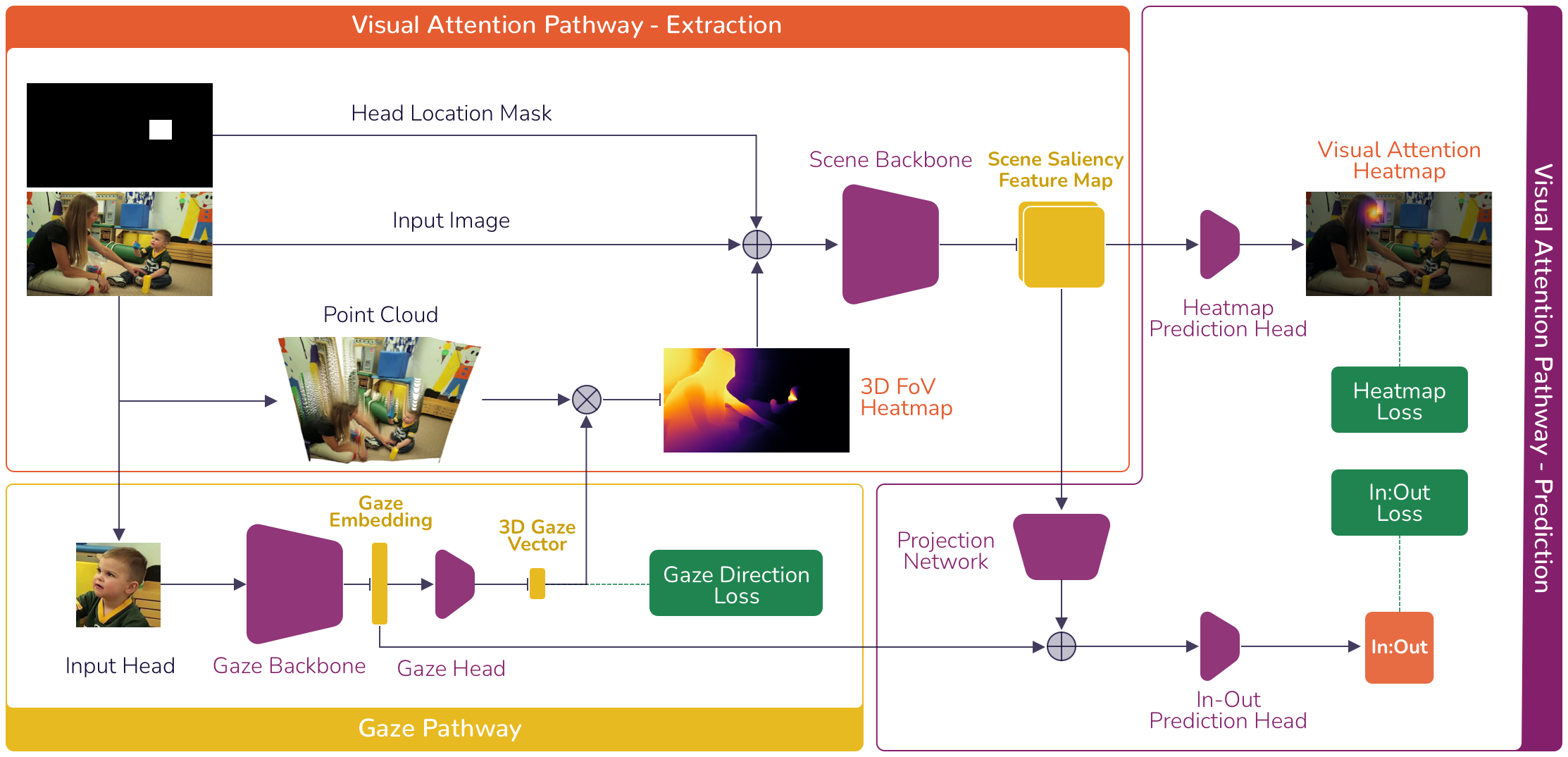}
    \vspace*{-2mm}
    \caption{
      Overview of our proposed architecture.
      The Gaze Network processes the head crop to predict a 3D gaze vector, which is then used with inferred point cloud to generate a heatmap of the 3D Field-of-view. 
      The Scene Pathway further combines this map with the image and a head location mask to predict a feature map highlighting salient items in the scene.
      This map is further used to predict on one hand the visual attention map $\heatmap_{p}$, and on the other hand,
      with the gaze embedding \gazeembedding, the in-vs-out gaze label.
    }
    \vspace*{-2mm}
    \label{fig:architecture}
\end{figure*}

Our network architecture is illustrated in Figure~\ref{fig:architecture}.
%
%
Similar to other methods \cite{chong2020dvisualtargetattention,Fang_2021_CVPR_DAM,gupta2022modular}, our architecture comprises two main pathways.
On one hand, the gaze pathways (\GP) aims at predicting the scene elements which are in the 3D Field of View (\tdfov) of the person, represented by the heatmap \pseudoheatmap.
To do so, it takes as input the image crop \headcrop of the person's head and predicts its 3D gaze direction $\predictedgaze$ as well as a gaze embedding $\gazeembedding$.
The gaze $\predictedgaze$ is then combined with the scene consistent 3D point cloud \pointcloud inferred from the image to generate the \tdfov heatmap. 

On the other hand, the visual attention pathway (\VAP) combines the image with \GP information
(the location of the head represented by the head mask \headmask, \pseudoheatmap and the gaze embedding $\gazeembedding$) to infer the in-out label $\inoutindicator_{p}$ (i.e. looking inside the frame or outside) and
the attention heatmap $\heatmap_{p}$.
We details these elements below.
However, given the importance of the scene structure representation, we first describe how the scene point cloud is obtained.

\vspace*{-1mm}

\subsection{Point cloud generation}
\label{sec:pointcloud}

\vspace*{-2mm}

To obtain our point cloud in the camera coordinate system $\pointcloud^{\cam} = \{ \threedpoint_i^{\cam} = (X_i^{\cam}, Y_i^{\cam}, Z_i^{\cam}) \}$
associated to the 2D pixels defined in the image plane $\twodpoint_i = (x_i, y_i)$, we need to know the scene depth 
as well as the intrinsic parameters of the camera.
As these are not available, we need to infer them from the data and make assumptions.

Regarding depth, we leverage the pre-trained model of~\cite{patakin2022depth} to predict the depth $Z^{\cam}_i$ of each pixel.
%
which generates geometrically consistent depth maps which are crucial for
doing a proper 3D analysis of the scene.
%
%
%
As to camera parameters, we make standard assumption: square pixels, no skew, and the principal point at the image center. 
The more important parameter is the focal length, which is required to avoid scene stretching. 
In this paper, we estimate it using the pre-trained model of~\cite{Wei2021leres}. 
As a result, denoting by $W$ and $H$ the image width and height,
we obtain the simplified projection equation\vspace*{-1mm}
\begin{equation}
    \begin{bmatrix}
    X_i^{\cam} \\
    Y_i^{\cam} \\
    Z_i^{\cam} \\
    \end{bmatrix} = \begin{bmatrix}
    f & 0 & W/2 \\
    0 & f & H/2 \\
    0 & 0 & 1\\
    \end{bmatrix}^{-1}.\begin{bmatrix}
    x_i.Z_i^{\cam} \\
    y_i.Z_i^{\cam} \\
    Z_i^{\cam} \\
    \end{bmatrix}
\end{equation}
enabling us to build our point cloud model $\pointcloud^{\cam}$.

Note that $\pointcloud^{\cam}$ is defined in the camera coordinate system.
However, as our aim is to evaluate the scene elements visible from the person's viewpoint,
we transform it in the local eye coordinate system \eyecoordinatesystem in which the gaze vector is predicted
(see next section), resulting in $\pointcloud^{\eye}$.
Following~\cite{kellnhofer2019gaze360}, the origin of \eyecoordinatesystem is defined by the eye location\footnote{In practice, for GazeFollow, we use the annotated eye location,
and for VideoAttentionTarget and ChildPlay datasets we use the center of the annotated bounding box as eye location.},
and the basis vectors $(E_x, E_y, E_z)$ are such that $E_z$ is the unit vector from the camera to the eye,
and $E_x$ and $E_y$ are in the plane perpendicular to $E_z$.

\vspace*{-1mm}

\subsection{Gaze pathway}
\label{sec:human-centric}
\label{sec:gazepathway}

\vspace*{-1.5mm}

This pathway comprises several steps to generate the \tdfov  heatmap \pseudoheatmap,
as described below.

\mypartitle{Gaze Prediction Network.}
Its aim is to predict the gaze direction $\predictedgaze$ defined in the local coordinate system \eyecoordinatesystem
associated with the head crop \headcrop \cite{kellnhofer2019gaze360}.
\eyecoordinatesystem is used rather than the camera coordinate system, 
as the gaze depends mainly on appearance (head pose and eyes) 
and not on the head location within the image.
This network is composed of a gaze prediction  backbone $\gazesubnetwork_b$ and of a gaze prediction head $\gazesubnetwork_h$.
The first one, $\gazesubnetwork_b$, is a ResNet-18~\cite{he2016deep_resnet} network that predicts the gaze embedding \gazeembedding from the  head crop \headcrop, 
while the second  is an MLP with 2 layers followed by tanh activation which transforms this gaze embedding into the unit 3D gaze vector prediction \predictedgaze:
\begin{equation}
    \gazeembedding = \gazesubnetwork_b(\headcrop) \mbox{ and }   \predictedgaze  = \gazesubnetwork_h(\gazeembedding).
\end{equation}

\mypartitle{\tdfov  heatmap \pseudoheatmap generation.}
Its goal is to highlight the scene parts  lying in the gaze direction of the person.
To do so, given  the point cloud $\pointcloud^{\eye}$ and the gaze prediction \predictedgaze,
we simply compute the cosine similarity \cosinesimilarity between $\predictedgaze$ and every point $\threedpoint_i^{\eye}$ in $\pointcloud^{\eye}$,
and further apply an exponential decay function for values with lower similarity to enhance
the scene parts which are more in the gaze focus: \vspace{-0.5em}
\begin{equation}
  \pseudoheatmap_i =
  \begin{cases} 
    \cosinesimilarity_i, & \text{if}\ \cosinesimilarity_i>0.9 \vspace{0.1cm}\\
    0.9 \times \frac{\exp(5 \times \cosinesimilarity_i)}{\exp(5\times0.9)}, & \text{otherwise.}
  \end{cases}
\vspace{-0.5em}
\end{equation}
and $\cosinesimilarity_i = \predictedgaze . \frac{\threedpoint_i^{\eye}}{||\threedpoint_i^{\eye}||}$.
%
%
Note that this formulation of the \tdfov is differentiable, allowing end-to-end training.

\vspace*{-1mm}

\subsection{Scene Pathway}

\vspace*{-1.5mm}

The scene pathway combines the scene information (the image \inputimage)
with the  \tdfov heatmap \pseudoheatmap of the person (to which we add the head location mask \headmask 
to better characterize the location and scale of the person in the scene)
and the gaze embedding \gazeembedding
to infer his attention (in-out indicator $\inoutindicator_{p}$ and visual attention heatmap $\heatmap_{p}$),
according to: \vspace{-0.5em}
\begin{equation}
  \featuremap = \featuresubnetwork([\inputimage, \pseudoheatmap, \headmask])
  \label{eq:featureextraction}
\end{equation}
\begin{equation}
  \heatmap_{p} = \regressionnetwork(\featuremap) \mbox{ and } \inoutindicator_{p} = \inoutnetwork([\sceneembedding, \gazeembedding]) \mbox{ with } \sceneembedding = \compressionnetwork(\featuremap)
\label{eq:predictions}
\vspace{-0.3em}
\end{equation}
which we explain below.

\mypartitle{Saliency feature extraction.}
The scene backbone network \featuresubnetwork is an encoder-decoder architecture  producing a set \featuremap of gaze saliency feature maps. 
The encoder is an EfficientNet-B1~\cite{tan2019efficientnet} network while the decoder is a Feature Pyramid Network (FPN)~\cite{lin2017fpn}.
The FPN contains skip connections that help retain high resolution spatial information which will improve gaze target localization.
The concatenation of inputs in Eq.~\ref{eq:featureextraction} can be considered as early fusion of the scene and gaze  information
and has been shown to give better performance compared to late fusion schemes, e.g.~\cite{gupta2022modular}.

\mypartitle{Attention prediction.}
It is summarized in Eq.~\ref{eq:predictions}, and comprises two parts.
The main one is the attention prediction head network \regressionnetwork which  process the feature maps \featuremap to predict
the gaze target heatmap $\heatmap_{p}$, whose maximum gives us the gaze target location.
It is a CNN block with 6 layers of dilated convolutions, and a 1x1 conv regression layer.

The second one is the in-out prediction head \inoutnetwork deciding whether the gaze target is within the frame.
It is  an MLP with 2 layers followed by a sigmoid activation that fuses the gaze embedding \gazeembedding with the scene embedding \sceneembedding to reach a decision.
The embedding \sceneembedding is derived from the gaze saliency feature maps \featuremap through the compression network \compressionnetwork
(a CNN block with 3 strided convolution layers followed by max pooling).

\vspace*{-1.5mm}

\subsection{Ground truth and loss definition}
\label{sec: ground-truth-loss}

\vspace*{-1.5mm}

\mypartitle{Heatmap GT.}
The gaze target location is encoded in a standard way ~\cite{chong2020dvisualtargetattention} as a GT heatmap $\heatmap_{gt}$ with a 2D isotropic gaussian centered
at the annotated gaze target location.

\mypartitle{3D Gaze Vector pseudo GT.}
While other methods only use the 2D information to drive the gaze pathway estimation,
we propose to use our geometrically consistent point cloud to define a pseudo 3D gaze direction ground truth.
Given the 2D gaze target GT, we obtain the corresponding 3D gaze point
$\threedpoint_{gaze}^{\eye}$ in the point cloud defined in the eye coordinate
system \eyecoordinatesystem (see Section~\ref{sec:pointcloud}),
and simply derive the 3D gaze unit vector accordingly as $\gtgaze = \frac{\threedpoint_{gaze}^{\eye}}{||\threedpoint_{gaze}^{\eye}||}$


\mypartitle{Loss definitions.}
Learning 
is driven by three losses:\\[-3.5mm]
\begin{equation}
    \loss = \losscoeff_{hm} \loss_{hm} + \losscoeff_{dir} \loss_{dir} + \losscoeff_{io} \loss_{io}.
\end{equation}
%
%
%
The first loss is the visual attention heatmap loss, defined as in other works as the L2 loss
between the predicted  and GT heatmaps: $\loss_{hm} = \|\heatmap_{p} - \heatmap_{gt}\|_2^2$.
%
The second loss is the in-out loss $\loss_{io}$, and is classically defined as the binary cross entropy between
the predicted $\inoutindicator_{p}$ and ground truth $\inoutindicator_{gt}$ in-vs-out of frame label.

Finally, to the contrary of previous works which relied only on 2D (in plane) gaze losses,
we propose in this work to introduce a 3D direction loss to drive the gaze pathway network.
It is defined so as to maximizes the cosine similarity between the prediction and the GT 3D gaze $\gtgaze$:
\begin{equation}
    \loss_{dir} = 1 - <\predictedgaze, \gtgaze>
\end{equation}
where $<a,b>$ denotes the inner product between $a$ and $b$.

\section{Experiments}
\label{sec:experiments}
\subsection{Experimental Protocol.}

\mypartitle{Implementation Details.}
The gaze network head $\gazesubnetwork_h$ is pre-trained on the Gaze360 dataset~\cite{kellnhofer2019gaze360} and processes the head crop at a resolution of $224\times224$. The Scene Pathway encoder is pre-trained on the ImageNet dataset~\cite{russakovsky2015imagenet} and processes the scene image at a resolution of $512\times288$. During the test phase, we maintain the original aspect ratio of the scene image and scale the longer side to $512$.

\mypartitle{Datasets.}
We train our models on 3 datasets - GazeFollow, ChildPlay and VideoAttentionTarget. More details about these datasets are provided in Section~\ref{sec:gaze-datasets}.

\mypartitle{Training.}
We train for 40 epochs on GazeFollow. Following the protocol of~\cite{chong2020dvisualtargetattention}, we fine-tune the model trained on GazeFollow for 20 epochs on VideoAttentionTarget. 
We  adopt the same protocol for ChildPlay.
We use the AdamW optimizer~\cite{loshchilov2018decoupled_adamw} and set the learning rate as $2.5e-4$ for training on GazeFollow, and as $2.5e-5$ for training on VideoAttentionTarget and ChildPlay. The loss coefficients are set as $\losscoeff_{hm}=100$, $\losscoeff_{dir}=0.1$ and $\losscoeff_{io}=1$.

\mypartitle{Validation.}
As GazeFollow and VideoAttentionTarget do not propose any validation set, we split a portion of the train set and use it for validation. Our GazeFollow val split contains 4499 instances, and our VideoAttentionTarget val split contains 6726 instances from 3 shows. The epoch with the best distance score on the validation set is used for testing.

\subsection{Metrics}

For evaluation, we use standard metric (AUC, Distance, AP) 
that we complement with a more semantic one: 
he precision of looking at heads (P.Head), as described below.

\mypartitle{AUC.}
The predicted gaze heatmap is compared against the binarized GT gaze heatmap. This is used to plot the TPR vs FPR curve. AUC is the area under this curve.

\mypartitle{Distance.}
The predicted gaze location is obtained from the arg max of the predicted gaze heatmap. The Distance is the L2 distance between the predicted and GT gaze location on a $1\times1$ image. When multiple annotations are available (ex. GazeFollow) we can compute the minimum and average distance statistics.

\mypartitle{Average Precision (AP).} It is used to compute the performance for in vs out of frame gaze classification. 

\begin{table}
\centering
\small
\begin{tabular}{C{0.12\linewidth} C{0.2\linewidth} C{0.1\linewidth} C{0.15\linewidth} C{0.15\linewidth}}
    \hline
    \textbf{\%Head} & \textbf{GazeFollow \cite{recasens2017following}} & \textbf{VAT \cite{chong2020dvisualtargetattention}} & \textbf{ChildPlay \textit{children}} & \textbf{ChildPlay \textit{adults}} \\
    \hline
     \textbf{All}  &  23.0 & 69.0 & 15.7 & 44.4\\
    \textbf{Multi}  &  30.6 &  71.0 & 16.9 & 44.6\\
    \hline
\end{tabular}
\caption{GT percentage of looking at head instances.
Statistics for all images (1st row) or  
images with at least 2 persons (2nd row).
}
\label{tab:gt-looking-head}
\vspace{-2mm}
\end{table}

\mypartitle{Looking at Heads (P.Head).}
The GT to compute this metric was obtained as follows.
We first run a robust and powerful pre-trained Yolo-v5~\cite{glenn_jocher_2022_7002879_yolov5} based head detector on images to get the head bounding boxes of everyone in the scene. 
We verified and further validated the obtained detections. 
To obtain the GT, we then check if the annotated gaze target of a person falls inside a detected head box. 
As the GazeFollow test set contains multiple annotation, 
we check that at least two annotations fall inside 
the same detected head box. 
We provide the GT statistics for our datasets in Table~\ref{tab:gt-looking-head}.

At evaluation time, we perform the same procedure for each  prediction to decide whether it is a gaze on a face. 
Finally, we compare the results with the GT, and compute the precision score.


\begin{table*}[t]
    \centering
    \small
    \vspace{-1em}
    \begin{tabular}{p{0.12\linewidth} C{0.04\linewidth} C{0.04\linewidth} C{0.04\linewidth} C{0.06\linewidth} | c c c c | c c c c}
        \hline
        & \multicolumn{4}{c}{Children} & \multicolumn{4}{c}{Adults} & \multicolumn{4}{c}{Full data} \\
        \textbf{Model} & \textbf{AUC$\uparrow$} & \textbf{Dist$\downarrow$} & \textbf{AP$\uparrow$} & \textbf{P.Head$\uparrow$} & \textbf{AUC$\uparrow$} & \textbf{Dist$\downarrow$} & \textbf{AP$\uparrow$} & \textbf{P.Head$\uparrow$} & \textbf{AUC$\uparrow$} & \textbf{Dist$\downarrow$} & \textbf{AP$\uparrow$} & \textbf{P.Head$\uparrow$}\\
        \hline
        \hline
        Gupta~\cite{gupta2022modular}$\dagger$ & 0.926 & 0.136 & - & 0.435 & 0.919 & 0.151 & - & 0.621 & 0.923 &  0.142 & - & 0.518\\
        Ours - 2D cone$\dagger$ & 0.929 & 0.125 & - & 0.472 & 0.934 & 0.131 & - & 0.664 & 0.931 & 0.127 & - & 0.567\\
        Ours$\dagger$ & 0.934 & 0.112 & - & 0.509 & 0.930 & \textcolor{red}{0.119} & - & 0.681 & 0.932 & 0.115 & - & 0.602\\
        \hline
        Gupta~\cite{gupta2022modular} & 0.923 & \textcolor{blue}{0.106} & \textcolor{blue}{0.980} & \textcolor{red}{0.648} & 0.914 & 0.123 & \textcolor{red}{0.987} & \textcolor{red}{0.731} & 0.919 & \textcolor{blue}{0.113} & \textcolor{blue}{0.983} & \textcolor{red}{0.694}\\
        Ours - 2D cone & 0.925 & 0.118 & 0.937 & 0.564 & 0.927 & 0.125 & 0.955 & \textcolor{blue}{0.717} & 0.926 & 0.121 & 0.944 & 0.644\\
        Ours & 0.939 & \textcolor{red}{0.098} & \textcolor{red}{0.989} & \textcolor{blue}{0.604} & 0.928 & \textcolor{blue}{0.121} & \textcolor{blue}{0.983} & 0.704 & 0.935 & \textcolor{red}{0.107} & \textcolor{red}{0.986} & \textcolor{blue}{0.663}\\
        \hline
        Human & - & - & - & - & - & - & - & - & 0.911 & 0.048 & 0.993 & - \\
        \hline
    \end{tabular}
    \caption{Results on the ChildPlay dataset. The best results are given in \textcolor{red}{red} and the second best results are given in \textcolor{blue}{blue}. $\dagger$ indicates that the model was not finetuned on ChildPlay. }
    \vspace{-1em}
    \label{tab:results-childplay}
\end{table*}

\subsection{Tested Models}

\mypartitle{ChildPlay.}
Other than our proposed model, we train the Image model of Gupta et al.~\cite{gupta2022modular} on our ChildPlay dataset. We also show results without any finetuning on ChildPlay, i.e. the models are only trained on GazeFollow.

\mypartitle{GazeFollow and VideoAttentionTarget.}
We train our proposed model on the GazeFollow and VideoAttentionTarget benchmarks. We also re-train the Image model of Gupta et al.~\cite{gupta2022modular} on VideoAttentionTarget following our new train and val splits. For state of the art, we compare results with the static model of Chong et al.~\cite{chong2020dvisualtargetattention}, as well as models using depth information and using the same head crop input: Fang et al.~\cite{Fang_2021_CVPR_DAM}, Hu et al.~\cite{hu2022we}, Gupta et al.~\cite{gupta2022modular}, Bao et al.~\cite{bao2022escnet} and Jin et al.~\cite{jin2022depth}. 

\mypartitle{Ablation: 3DFoV vs 2D cone.}
To see the benefit of using an explicit 3D saliency model, 
we compare our approach to using a standard 2D gaze cone (similarly to \cite{gupta2022modular}) on the VideoAttentionTarget and ChildPlay datasets. 
Here the 2D gaze cone is derived by computing 
the similarity of the projected 3D gaze vector and 
the 2D scene locations  (no decay factor). 

\subsection{Results}

\begin{table}[t]
    \centering
    \small
    \begin{tabular}{p{0.25\linewidth} C{0.09\linewidth} C{0.13\linewidth} C{0.13\linewidth} C{0.14\linewidth}}
        \hline
        \textbf{A: Model} & \textbf{AUC$\uparrow$} & \textbf{Avg.Dist$\downarrow$} & \textbf{Min.Dist$\downarrow$} & \textbf{P.Head$\uparrow$}\\
        \hline
        \hline
        Fang~\cite{Fang_2021_CVPR_DAM} & 0.922 & 0.124 & 0.067 & -\\
        Hu~\cite{hu2022we} & - & 0.135 & 0.075 & -\\
        Bao~\cite{bao2022escnet} & 0.928 & \textcolor{blue}{0.122} & - & -\\
        Jin~\cite{jin2022depth} & 0.920 & \textcolor{red}{0.118} & \textcolor{blue}{0.063} & -\\
        Chong~\cite{chong2020dvisualtargetattention} & 0.921 & 0.137 & 0.077 & 0.708\\
        Gupta~\cite{gupta2022modular} & 0.933 & 0.134 & 0.071 & 0.750\\
        \hline
        Ours - 2D cone* & 0.939 & \textcolor{blue}{0.122} & \textcolor{red}{0.062} & \textcolor{red}{0.762}\\
        Ours* & 0.936 & 0.125 & 0.064 & \textcolor{blue}{0.760} \\
        \hline
        Human & 0.924 & 0.096 & 0.040 & -\\
        \hline
    \end{tabular}

    \vspace{1em}
    \begin{tabular}{p{0.25\linewidth} C{0.1\linewidth} C{0.12\linewidth} C{0.12\linewidth} C{0.15\linewidth}}
        \hline
        \textbf{B: Model} & \textbf{AUC$\uparrow$} & \textbf{Dist$\downarrow$} & \textbf{AP$\uparrow$} & \textbf{P.Head$\uparrow$}\\
        \hline
        \hline
         Gupta~\cite{gupta2022modular}$\dagger$ & 0.907 & 0.137 & - & 0.887\\
         Ours - 2D cone$\dagger$ & 0.915 & 0.128 & - & 0.894\\
         Ours$\dagger$ & 0.911 & 0.123 & - & 0.900\\
        \hline
        Fang~\cite{Fang_2021_CVPR_DAM} & 0.878 & 0.124 & \textcolor{blue}{0.872} & -\\
        Bao~\cite{bao2022escnet} & 0.885 & \textcolor{blue}{0.120} & 0.869 & -\\
        Jin~\cite{jin2022depth} & 0.898 & \textcolor{red}{0.109} & \textcolor{red}{0.897} & -\\
        Chong~\cite{chong2020dvisualtargetattention} &  0.854 & 0.147 & 0.848 & -\\
        Gupta~\cite{gupta2022modular}* & 0.897 & 0.134 & 0.864 & \textcolor{red}{0.903}\\
        \hline
        Ours - 2D cone* & 0.909 & \textcolor{blue}{0.120} & 0.856 & 0.892\\
        Ours* & 0.914 & \textcolor{red}{0.109} & 0.834 & \textcolor{blue}{0.902}\\
        \hline
        Human & 0.921 & 0.051 & 0.925 & -\\
        \hline
    \end{tabular}
    \caption{Results on GazeFollow (A) and VideoAttentionTarget (B) with the best results in \textcolor{red}{red} and 
    second best results in \textcolor{blue}{blue}. * indicates that the model follows a proper protocol, using a validation split to select the model. $\dagger$ indicates that the model was not fine-tuned on VAT.}
    \label{tab:results-videoatt}
    \vspace{-1.5em}
    \label{tab:results-gazefollow-videoatt}
\end{table}

\mypartitle{GazeFollow and VideoAttentionTarget (VAT).}
Our results on GazeFollow and VideoAttentionTarget are given in Table~\ref{tab:results-gazefollow-videoatt}. 
As it can be seen, our model achieves high results.  
Compared to the state-of-the-art, it is in par 
with the best method \cite{jin2022depth}, 
which also used depth but without modeling 
an explicit 3DFoV, and may not use a validation set 
for evaluation. 
%
Indeed, on GazeFollow, although the \AvgDist is slightly worse, 
the Min.Dist metric is the same, 
and on VAT dataset both methods perform equally for Dist (0.109). 
Compared to \cite{gupta2022modular} which follows the same protocol, our method performs much better.

Looking at the P.Head metric, we can notice that performance is in general quite high, esp. on the VAT dataset that has a large bias towards looking at heads (Table~\ref{tab:gt-looking-head}), 
so the true positives dominate the false positives. 

\mypartitle{ChildPlay.}
%
Our results on ChildPlay are given in Table~\ref{tab:results-childplay}.
As it can be seen, our model shows much better cross-dataset generalization performance compared to our model with a 2D gaze cone and the model of Gupta et al.~\cite{gupta2022modular}.
We see a general improvement in performance for all models after fine-tuning. 
Although the benefit is lower for our model, 
it is still the best. 
%
Ultimately and interestingly, the gap in performance compared to  human performance  suggests a large potential for improvement.

\mypartitle{Children vs Adults.}
Looking more in details, ChildPlay results show 
that the distance scores are slightly better for 
 children compared to adults. 
 However, this can mainly be attributed to the fact 
 that  gaze targets  are on average closer 
 to the child than to the adult (see Fig~\ref{fig:childplay-stats}), a point also reported by 
 Tu et al.~\cite{tu2022end} 
 who showed that  gaze target prediction models 
 have lower performance (using a distance metric) 
 for targets further away. 

This contrasts with the P.Head metric, 
which shows that in this case, the performance is significantly 
lower for children than for adults (18.7\% lower). 
This validates our hypothesis that different performance metrics are needed to fully assess models, and that models trained on existing datasets suffer when tested on children. 
This last point is corroborated by the fact that after fine-tuning, all models have a much larger improvement for children compared to adults whether for the distance or P.Head metrics, highlighting the importance of training the model with children data. 

\mypartitle{3DFoV vs 2D cone saliency.}
Results show that our model (Ours) with an explicit 3DFoV 
performs on par or much better than a  model relying only a 2D saliency cone (Ours-2D cone) 
On GazeFollow, results are very slightly worse, 
which can be due to the fact that GazeFollow 
contains relatively simple scenes where depth is less important, 
and with a bias towards people being in the foreground 
(in the vast majority of a images, only 
the gaze of the person with the largest face in the scene is annotated).
However, on VideoAttentionTarget and ChildPlay, our model with 3DFoV demonstrates much better cross-dataset generalization, 
and remains significantly better after fine-tuning. 
 All this highlights the importance of depth information
 and the interest of using a geometrically consistent 
 3DFoV method.



\vspace*{-1mm}
\section{Conclusion}
\label{sec:conclusions}

\vspace*{-2mm}

In this paper we proposed a new dataset of children playing and interacting with adults. Our dataset has rich gaze annotations which is of interest for analysis of child gaze behaviour and gaze target prediction generally. 
We also proposed a new model that uses depth information to construct a geometrically grounded 3D field of view of a person.
Our models achieve state of the art results on public benchmarks and ChildPlay.
In particular, experiments indicate  that training on ChildPlay can yield performance improvements for child gaze prediction, 
and that using semantic metrics (looking at faces) is useful to 
further characterize gaze models.
In the future we will supplement ChildPlay in the future with other layers of annotations (e.g. human-human-object interaction labels) and we encourage the research community to do the same.

{\small
\bibliographystyle{ieee_fullname}
\bibliography{ChildPlayGaze}

\begin{thebibliography}{10}\itemsep=-1pt

\bibitem{Anzalone:PRL:2019}
Salvatore~Maria Anzalone, Jean Xavier, Sofiane Boucenna, Lucia Billeci, Antonio
  Narzisi, Filippo Muratori, David Cohen, and Mohamed Chetouani.
\newblock {Quantifying patterns of joint attention during human-robot
  interactions: An application for autism spectrum disorder assessment}.
\newblock {\em Pattern Recognition Letters}, 118:42--50, 2019.

\bibitem{bao2022escnet}
Jun Bao, Buyu Liu, and Jun Yu.
\newblock Escnet: Gaze target detection with the understanding of 3d scenes.
\newblock In {\em Proceedings of the IEEE/CVF Conference on Computer Vision and
  Pattern Recognition}, pages 14126--14135, 2022.

\bibitem{belkada2021pedestrians}
Younes Belkada, Lorenzo Bertoni, Romain Caristan, Taylor Mordan, and Alexandre
  Alahi.
\newblock Do pedestrians pay attention? eye contact detection in the wild,
  2021.

\bibitem{billing2020dream}
Erik Billing, Tony Belpaeme, Haibin Cai, Hoang-Long Cao, Anamaria Ciocan,
  Cristina Costescu, Daniel David, Robert Homewood, Daniel Hernandez~Garcia,
  Pablo G{\'o}mez~Esteban, et~al.
\newblock The dream dataset: Supporting a data-driven study of autism spectrum
  disorder and robot enhanced therapy.
\newblock {\em PloS one}, 15(8):e0236939, 2020.

\bibitem{cheng2022masked}
Bowen Cheng, Ishan Misra, Alexander~G Schwing, Alexander Kirillov, and Rohit
  Girdhar.
\newblock Masked-attention mask transformer for universal image segmentation.
\newblock In {\em Proceedings of the IEEE/CVF Conference on Computer Vision and
  Pattern Recognition}, pages 1290--1299, 2022.

\bibitem{Chong:IMWU:2017}
Eunji Chong, Katha Chanda, Zhefan Ye, Audrey Southerland, Nataniel Ruiz,
  Rebecca~M. Jones, Agata Rozga, and James~M. Rehg.
\newblock {Detecting Gaze Towards Eyes in Natural Social Interactions and Its
  Use in Child Assessment}.
\newblock {\em Proceedings of the ACM on Interactive, Mobile, Wearable and
  Ubiquitous Technologies}, 1(3):1--20, 2017.

\bibitem{chong2017detecting}
Eunji Chong, Katha Chanda, Zhefan Ye, Audrey Southerland, Nataniel Ruiz,
  Rebecca~M Jones, Agata Rozga, and James~M Rehg.
\newblock Detecting gaze towards eyes in natural social interactions and its
  use in child assessment.
\newblock {\em Proceedings of the ACM on Interactive, Mobile, Wearable and
  Ubiquitous Technologies}, 1(3):1--20, 2017.

\bibitem{chong2018connecting}
Eunji Chong, Nataniel Ruiz, Yongxin Wang, Yun Zhang, Agata Rozga, and James~M
  Rehg.
\newblock Connecting gaze, scene, and attention: Generalized attention
  estimation via joint modeling of gaze and scene saliency.
\newblock In {\em Proceedings of the European conference on computer vision
  (ECCV)}, pages 383--398, 2018.

\bibitem{chong2020dvisualtargetattention}
Eunji Chong, Yongxin Wang, Nataniel Ruiz, and James~M Rehg.
\newblock Detecting attended visual targets in video.
\newblock In {\em Proceedings of the IEEE/CVF Conference on Computer Vision and
  Pattern Recognition}, pages 5396--5406, 2020.

\bibitem{de2020computer}
Ryan Anthony~J de Belen, Tomasz Bednarz, Arcot Sowmya, and Dennis Del~Favero.
\newblock Computer vision in autism spectrum disorder research: a systematic
  review of published studies from 2009 to 2019.
\newblock {\em Translational psychiatry}, 10(1):1--20, 2020.

\bibitem{edition2013_dsm5}
Fifth Edition et~al.
\newblock Diagnostic and statistical manual of mental disorders.
\newblock {\em American Psychiatric Association}, 21:591--643, 2013.

\bibitem{fan2018inferring_videocoatt}
Lifeng Fan, Yixin Chen, Ping Wei, Wenguan Wang, and Song-Chun Zhu.
\newblock Inferring shared attention in social scene videos.
\newblock In {\em Proceedings of the IEEE conference on computer vision and
  pattern recognition}, pages 6460--6468, 2018.

\bibitem{Fang_2021_CVPR_DAM}
Yi Fang, Jiapeng Tang, Wang Shen, Wei Shen, Xiao Gu, Li Song, and Guangtao
  Zhai.
\newblock Dual attention guided gaze target detection in the wild.
\newblock In {\em Proceedings of the IEEE/CVF Conference on Computer Vision and
  Pattern Recognition (CVPR)}, pages 11390--11399, June 2021.

\bibitem{franchak2016free}
John~M Franchak, David~J Heeger, Uri Hasson, and Karen~E Adolph.
\newblock Free viewing gaze behavior in infants and adults.
\newblock {\em Infancy}, 21(3):262--287, 2016.

\bibitem{geiger2013kitti}
Andreas Geiger, Philip Lenz, Christoph Stiller, and Raquel Urtasun.
\newblock Vision meets robotics: The kitti dataset.
\newblock {\em The International Journal of Robotics Research},
  32(11):1231--1237, 2013.

\bibitem{guan2020_pose}
Jian Guan, Liming Yin, Jianguo Sun, Shuhan Qi, Xuan Wang, and Qing Liao.
\newblock Enhanced gaze following via object detection and human pose
  estimation.
\newblock In {\em International Conference on Multimedia Modeling}, pages
  502--513. Springer, 2020.

\bibitem{guan2020enhanced}
Jian Guan, Liming Yin, Jianguo Sun, Shuhan Qi, Xuan Wang, and Qing Liao.
\newblock Enhanced gaze following via object detection and human pose
  estimation.
\newblock In {\em International Conference on Multimedia Modeling}, pages
  502--513. Springer, 2020.

\bibitem{gupta2022modular}
Anshul Gupta, Samy Tafasca, and Jean-Marc Odobez.
\newblock A modular multimodal architecture for gaze target prediction:
  Application to privacy-sensitive settings.
\newblock In {\em Proceedings of the IEEE/CVF Conference on Computer Vision and
  Pattern Recognition Workshops}, pages 5041--5050, 2022.

\bibitem{Hashemi:TAC:2018}
Jordan Hashemi, Geraldine Dawson, Kimberly~L.H. Carpenter, Kathleen Campbell,
  Qiang Qiu, Steven Espinosa, Samuel Marsan, Jeffery~P. Baker, Helen~L. Egger,
  and Guillermo Sapiro.
\newblock {Computer Vision Analysis for Quantification of Autism Risk
  Behaviors}.
\newblock {\em IEEE Transactions on Affective Computing}, 3045(AUGUST):1--12,
  2018.

\bibitem{he2016deep_resnet}
Kaiming He, Xiangyu Zhang, Shaoqing Ren, and Jian Sun.
\newblock Deep residual learning for image recognition.
\newblock In {\em Proceedings of the IEEE conference on computer vision and
  pattern recognition}, pages 770--778, 2016.

\bibitem{hesse2018computer}
Nikolas Hesse, Christoph Bodensteiner, Michael Arens, Ulrich~G Hofmann, Raphael
  Weinberger, and A Sebastian~Schroeder.
\newblock Computer vision for medical infant motion analysis: State of the art
  and rgb-d data set.
\newblock In {\em Proceedings of the European Conference on Computer Vision
  (ECCV) Workshops}, pages 0--0, 2018.

\bibitem{hu2022we}
Zhengxi Hu, Dingye Yang, Shilei Cheng, Lei Zhou, Shichao Wu, and Jingtai Liu.
\newblock We know where they are looking at from the rgb-d camera: Gaze
  following in 3d.
\newblock {\em IEEE Transactions on Instrumentation and Measurement}, 2022.

\bibitem{huang2021invariant}
Xiaofei Huang, Nihang Fu, Shuangjun Liu, and Sarah Ostadabbas.
\newblock Invariant representation learning for infant pose estimation with
  small data.
\newblock In {\em 2021 16th IEEE International Conference on Automatic Face and
  Gesture Recognition (FG 2021)}, pages 1--8. IEEE, 2021.

\bibitem{jin2021multi}
Tianlei Jin, Zheyuan Lin, Shiqiang Zhu, Wen Wang, and Shunda Hu.
\newblock Multi-person gaze-following with numerical coordinate regression.
\newblock In {\em 2021 16th IEEE International Conference on Automatic Face and
  Gesture Recognition (FG 2021)}, pages 01--08. IEEE, 2021.

\bibitem{jin2022depth}
Tianlei Jin, Qizhi Yu, Shiqiang Zhu, Zheyuan Lin, Jie Ren, Yuanhai Zhou, and
  Wei Song.
\newblock Depth-aware gaze-following via auxiliary networks for robotics.
\newblock {\em Engineering Applications of Artificial Intelligence},
  113:104924, 2022.

\bibitem{glenn_jocher_2022_7002879_yolov5}
Glenn Jocher, Ayush Chaurasia, Alex Stoken, Jirka Borovec, NanoCode012, Yonghye
  Kwon, TaoXie, Kalen Michael, Jiacong Fang, imyhxy, Lorna, Colin Wong, Zeng
  Yifu, Abhiram V, Diego Montes, Zhiqiang Wang, Cristi Fati, Jebastin Nadar,
  Laughing, UnglvKitDe, tkianai, yxNONG, Piotr Skalski, Adam Hogan, Max
  Strobel, Mrinal Jain, Lorenzo Mammana, and xylieong.
\newblock {ultralytics/yolov5: v6.2 - YOLOv5 Classification Models, Apple M1,
  Reproducibility, ClearML and Deci.ai integrations}, Aug. 2022.

\bibitem{kellnhofer2019gaze360}
Petr Kellnhofer, Adria Recasens, Simon Stent, Wojciech Matusik, and Antonio
  Torralba.
\newblock Gaze360: Physically unconstrained gaze estimation in the wild.
\newblock In {\em Proceedings of the IEEE/CVF International Conference on
  Computer Vision}, pages 6912--6921, 2019.

\bibitem{labelboxwebsite}
Labelbox.
\newblock Labelbox, online, 2022.

\bibitem{li2016modified}
Beibin Li, Quan Wang, Erin Barney, Logan Hart, Carla Wall, Katarzyna Chawarska,
  Irati~Saez de Urabain, Timothy~J Smith, and Frederick Shic.
\newblock Modified dbscan algorithm on oculomotor fixation identification.
\newblock In {\em Proceedings of the Ninth Biennial ACM Symposium on Eye
  Tracking Research \& Applications}, pages 337--338, 2016.

\bibitem{li2018megadepth}
Zhengqi Li and Noah Snavely.
\newblock Megadepth: Learning single-view depth prediction from internet
  photos.
\newblock In {\em Proceedings of the IEEE conference on computer vision and
  pattern recognition}, pages 2041--2050, 2018.

\bibitem{lian2018believe}
Dongze Lian, Zehao Yu, and Shenghua Gao.
\newblock Believe it or not, we know what you are looking at!
\newblock In {\em Asian Conference on Computer Vision}, pages 35--50. Springer,
  2018.

\bibitem{lin2017fpn}
Tsung-Yi Lin, Piotr Doll{\'a}r, Ross Girshick, Kaiming He, Bharath Hariharan,
  and Serge Belongie.
\newblock Feature pyramid networks for object detection.
\newblock In {\em Proceedings of the IEEE conference on computer vision and
  pattern recognition}, pages 2117--2125, 2017.

\bibitem{loshchilov2018decoupled_adamw}
Ilya Loshchilov and Frank Hutter.
\newblock Decoupled weight decay regularization.
\newblock In {\em International Conference on Learning Representations}, 2018.

\bibitem{marin2019laeo}
Manuel~J Marin-Jimenez, Vicky Kalogeiton, Pablo Medina-Suarez, and Andrew
  Zisserman.
\newblock Laeo-net: revisiting people looking at each other in videos.
\newblock In {\em Proceedings of the IEEE/CVF Conference on Computer Vision and
  Pattern Recognition}, pages 3477--3485, 2019.

\bibitem{migliorelli2020babypose}
Lucia Migliorelli, Sara Moccia, Rocco Pietrini, Virgilio~Paolo Carnielli, and
  Emanuele Frontoni.
\newblock The babypose dataset.
\newblock {\em Data in brief}, 33:106329, 2020.

\bibitem{mundy1990longitudinal}
Peter Mundy, Marian Sigman, and Connie Kasari.
\newblock A longitudinal study of joint attention and language development in
  autistic children.
\newblock {\em Journal of Autism and developmental Disorders}, 20(1):115--128,
  1990.

\bibitem{nan2021predicting}
Zhixiong Nan, Jingjing Jiang, Xiaofeng Gao, Sanping Zhou, Weiliang Zuo, Ping
  Wei, and Nanning Zheng.
\newblock Predicting task-driven attention via integrating bottom-up stimulus
  and top-down guidance.
\newblock {\em IEEE Transactions on Image Processing}, 30:8293--8305, 2021.

\bibitem{patakin2022depth}
Nikolay Patakin, Anna Vorontsova, Mikhail Artemyev, and Anton Konushin.
\newblock Single-stage 3d geometry-preserving depth estimation model training
  on dataset mixtures with uncalibrated stereo data.
\newblock In {\em Proceedings of the IEEE/CVF Conference on Computer Vision and
  Pattern Recognition}, pages 1705--1714, 2022.

\bibitem{rajagopalan2013self}
Shyam Rajagopalan, Abhinav Dhall, and Roland Goecke.
\newblock Self-stimulatory behaviours in the wild for autism diagnosis.
\newblock In {\em Proceedings of the IEEE International Conference on Computer
  Vision Workshops}, pages 755--761, 2013.

\bibitem{ranftl2019midas}
Ren\'{e} Ranftl, Katrin Lasinger, David Hafner, Konrad Schindler, and Vladlen
  Koltun.
\newblock Towards robust monocular depth estimation: Mixing datasets for
  zero-shot cross-dataset transfer.
\newblock {\em IEEE Transactions on Pattern Analysis and Machine Intelligence
  (TPAMI)}, 2020.

\bibitem{nips15_recasens}
Adria Recasens$^*$, Aditya Khosla$^*$, Carl Vondrick, and Antonio Torralba.
\newblock Where are they looking?
\newblock In {\em Advances in Neural Information Processing Systems (NIPS)},
  2015.
\newblock $^*$ indicates equal contribution.

\bibitem{recasens2017following}
Adria Recasens, Carl Vondrick, Aditya Khosla, and Antonio Torralba.
\newblock Following gaze in video.
\newblock In {\em Proceedings of the IEEE International Conference on Computer
  Vision}, pages 1435--1443, 2017.

\bibitem{rehg2013decoding}
James Rehg, Gregory Abowd, Agata Rozga, Mario Romero, Mark Clements, Stan
  Sclaroff, Irfan Essa, O Ousley, Yin Li, Chanho Kim, et~al.
\newblock Decoding children's social behavior.
\newblock In {\em Proceedings of the IEEE conference on computer vision and
  pattern recognition}, pages 3414--3421, 2013.

\bibitem{Rehg:CVPR:2013}
J~M Rehg, G~D Abowd, A Rozga, M Romero, M~A Clements, S Sclaroff, I Essa, O~Y
  Ousley, Yin Li, Chanho Kim, H Rao, J~C Kim, L~L Presti, Jianming Zhang, D
  Lantsman, J Bidwell, and Zhefan Ye.
\newblock {Decoding Children's Social Behavior}.
\newblock In {\em Computer Vision and Pattern Recognition (CVPR)}, jun 2013.

\bibitem{rihawi20173d}
Omar Rihawi, Djamal Merad, and Jean-Luc Damoiseaux.
\newblock 3d-ad: 3d-autism dataset for repetitive behaviours with kinect
  sensor.
\newblock In {\em 2017 14th IEEE International Conference on Advanced Video and
  Signal Based Surveillance (AVSS)}, pages 1--6. IEEE, 2017.

\bibitem{Rudovic2018}
Ognjen Rudovic, Jaeryoung Lee, Miles Dai, Bj{\"{o}}rn Schuller, and Rosalind~W.
  Picard.
\newblock {Personalized machine learning for robot perception of affect and
  engagement in autism therapy}.
\newblock {\em Science Robotics}, 3(19):1--12, 2018.

\bibitem{russakovsky2015imagenet}
Olga Russakovsky, Jia Deng, Hao Su, Jonathan Krause, Sanjeev Satheesh, Sean Ma,
  Zhiheng Huang, Andrej Karpathy, Aditya Khosla, Michael Bernstein, et~al.
\newblock Imagenet large scale visual recognition challenge.
\newblock {\em International journal of computer vision}, 115(3):211--252,
  2015.

\bibitem{sciortino2017estimation}
Giuseppa Sciortino, Giovanni~Maria Farinella, Sebastiano Battiato, Marco Leo,
  and Cosimo Distante.
\newblock On the estimation of children’s poses.
\newblock In {\em International conference on image analysis and processing},
  pages 410--421. Springer, 2017.

\bibitem{senju2009atypical}
Atsushi Senju and Mark~H Johnson.
\newblock Atypical eye contact in autism: models, mechanisms and development.
\newblock {\em Neuroscience \& Biobehavioral Reviews}, 33(8):1204--1214, 2009.

\bibitem{shic2011limited}
Frederick Shic, Jessica Bradshaw, Ami Klin, Brian Scassellati, and Katarzyna
  Chawarska.
\newblock Limited activity monitoring in toddlers with autism spectrum
  disorder.
\newblock {\em Brain research}, 1380:246--254, 2011.

\bibitem{syeda2017visual}
Uzma~Haque Syeda, Ziaul Zafar, Zishan~Zahidul Islam, Syed~Mahir Tazwar,
  Miftahul~Jannat Rasna, Koichi Kise, and Md~Atiqur~Rahman Ahad.
\newblock Visual face scanning and emotion perception analysis between autistic
  and typically developing children.
\newblock In {\em Proceedings of the 2017 acm international joint conference on
  pervasive and ubiquitous computing and proceedings of the 2017 acm
  international symposium on wearable computers}, pages 844--853, 2017.

\bibitem{tan2019efficientnet}
Mingxing Tan and Quoc Le.
\newblock Efficientnet: Rethinking model scaling for convolutional neural
  networks.
\newblock In {\em International Conference on Machine Learning}, pages
  6105--6114. PMLR, 2019.

\bibitem{tomas2021goo}
Henri Tomas, Marcus Reyes, Raimarc Dionido, Mark Ty, Jonric Mirando, Joel
  Casimiro, Rowel Atienza, and Richard Guinto.
\newblock Goo: A dataset for gaze object prediction in retail environments.
\newblock In {\em Proceedings of the IEEE/CVF Conference on Computer Vision and
  Pattern Recognition}, pages 3125--3133, 2021.

\bibitem{tu2022end}
Danyang Tu, Xiongkuo Min, Huiyu Duan, Guodong Guo, Guangtao Zhai, and Wei Shen.
\newblock End-to-end human-gaze-target detection with transformers.
\newblock {\em arXiv preprint arXiv:2203.10433}, 2022.

\bibitem{wang2022gatector}
Binglu Wang, Tao Hu, Baoshan Li, Xiaojuan Chen, and Zhijie Zhang.
\newblock Gatector: A unified framework for gaze object prediction.
\newblock In {\em Proceedings of the IEEE/CVF Conference on Computer Vision and
  Pattern Recognition}, pages 19588--19597, 2022.

\bibitem{yin2020diversedepth}
Wei Yin, Xinlong Wang, Chunhua Shen, Yifan Liu, Zhi Tian, Songcen Xu, Changming
  Sun, and Dou Renyin.
\newblock Diversedepth: Affine-invariant depth prediction using diverse data.
\newblock {\em arXiv preprint arXiv:2002.00569}, 2020.

\bibitem{Wei2021leres}
Wei Yin, Jianming Zhang, Oliver Wang, Simon Niklaus, Long Mai, Simon Chen, and
  Chunhua Shen.
\newblock Learning to recover 3d scene shape from a single image.
\newblock In {\em Proceedings of the IEEE/CVF Conference on Computer Vision and
  Pattern Recognition}, 2021.

\bibitem{zhao2020learning}
Hao Zhao, Ming Lu, Anbang Yao, Yurong Chen, and Li Zhang.
\newblock Learning to draw sight lines.
\newblock {\em International Journal of Computer Vision}, 128(5):1076--1100,
  2020.

\end{thebibliography}
}

\clearpage
\section{Supplementary}
\label{sec:supplementary}

\subsection{More information on ChildPlay}

\mypartitle{Gaze Classes.}
ChildPlay is annotated with 7 non-overlapping gaze classes to enable high quality gaze annotations. These are defined as follows:

\begin{itemize}
\item inside-frame: when the gaze target is located within the frame and is visible;
\item outside-frame: the gaze target is  outside the frame;
\item gaze-shift: when the person shifts attention from one location to the next during at least two frames.
  In case of interest, shorter shifts (i.e. saccades)
  can be recovered by identifying sudden changes in gaze points
  that are annotated as inside-frame;
  \item occluded: the 2D gaze target is  within the frame but is totally occluded (hence cannot be annotated);
  \item uncertain: the gaze target cannot be determined confidently (lack of salient elements in the gaze direction,
  several possible targets);
       %
  %
\item eyes-closed: used in rare cases where a child closes their eyes (e.g. during hide-and-seek);
  %
  %
\item not-annotated: none of the options above is applicable.
\end{itemize}

\mypartitle{Semantics.}
We compare the semantics of the gaze targets for ChildPlay and VideoAttentionTarget in Table~\ref{tab: gaze-semantic-class}. Our ChildPlay dataset is far more balanced\footnote{
After manual inspection, we found that most of the  \texttt{not-detected} instances in ChildPlay correspond
to objects that were not detected by the segmentation, and  which would fall into
the \texttt{things-other} category.
},
while also having 50\% more frames and twice as much scene variety.

\subsection{More Children Datasets}

One of the major motivations behind building datasets of children is the study of neurodevelopmental disorders exhibiting symptoms in humans from an early age. For this reason, many benchmarks studied in the literature cover topics such as motor control, brain imaging, emotions, speech, and social interactions. Nevertheless, most of them are ultimately never shared due to privacy considerations and ethics regulations \cite{de2020computer}. We previously listed some of the children datasets directly related to autism behaviors, in this section, we cover a few publicly available ones that feature pose annotations. Since the  body proportions of humans change significantly from birth to adulthood \cite{sciortino2017estimation}, it is important for younger age groups to be well represented in research benchmarks, particularly for applications targeting them. Table \ref{tab: children-datasets} summarizes the notable ones.



\begin{table}[t]
\footnotesize
\centering
\begin{tabular}{
p{0.26\linewidth}p{0.12\linewidth}p{0.12\linewidth}p{0.1\linewidth}p{0.12\linewidth}
}
    \toprule
    \textbf{Dataset} & \textbf{things-person} & \textbf{things-other} & \textbf{stuff} & \textbf{not-detected}\\
    \addlinespace[2pt]
    \midrule
    
    VideoAttention \cite{chong2020dvisualtargetattention} & $80.85\%$ & $8.05\%$ & $3.60\%$ & $7.50\%$ \\
    
    \midrule
    
    ChildPlay & $45.19\%$ & $18.66\%$ & $12.62\%$ & $23.53\%$ \\
    
    \bottomrule
\end{tabular}
\caption{Comparison of gaze target semantic class between ChildPlay and VideoAttentionTarget.
  Numbers were obtained by running a panoptic segmentation model \cite{cheng2022masked} on images
  and retrieving the semantic class of each annotated gaze point.
}
\label{tab: gaze-semantic-class}
\vspace*{-4mm}
\end{table}

\subsection{Point Cloud Comparison}

\mypartitle{Monocular Depth Estimation.}
Depth datasets can be put under three categories:
\begin{itemize}
    \item Absolute Depth: These datasets provide the absolute depth of the scene. The data is recorded using sensors such as LiDARS, time of flight cameras etc. ex. KITTI~\cite{geiger2013kitti}
    
    \item Up to Scale (UTS) Depth: These datasets provide the depth of the scene up to an unknown scale $C_1$. The absolute depth $d*$ can be recovered from UTS depth $d$ as $d*^{-1} = C_1.d^{-1}$. ex. Megadepth~\cite{li2018megadepth}
    
    \item Up to Shift and Scale (UTSS) Depth: These datasets provide the disparity of scene. They are obtained from stereo movies and photos by computing the optical flow. The absolute depth can be recovered from the disparity $D$ as $d*^{-1} = C_1.(D + C_2)$. $C_2$, also known as shift, depends on the camera parameters and is crucial for reconstructing geometry preserving point clouds. However, the shift is typically unknown. ex. MiDaS~\cite{ranftl2019midas}
\end{itemize}

Recent methods for monocular depth estimation~\cite{ranftl2019midas}\cite{yin2020diversedepth} have leveraged UTSS depth data due to it's high diversity, and shown better generalization when tested on unseen datasets. However, they can only predict UTSS depth so the reconstructed point clouds are not geometry preserving. Hence, methods for gaze target prediction that use these algorithms rely on course matching~\cite{Fang_2021_CVPR_DAM} or attempt to correct the point cloud based on prior assumptions~\cite{bao2022escnet}.

We study two recent methods for monocular depth estimation that aim to generate geometry-preserving point clouds while still leveraging UTSS data. Wei et al.~\cite{Wei2021leres} predict UTSS depth and use it to construct a (distorted) point cloud. A point cloud module then recovers the shift factor from the distorted point cloud. On the other hand, Patakin et al.~\cite{patakin2022depth} train on a mix of absolute, UTS and UTSS depth data. The absolute and UTS depth data provide supervision such that the algorithm predicts UTS depth.

\begin{table*}[th]
\centering
\begin{tabular}{
p{0.2\linewidth}p{0.1\linewidth}p{0.25\linewidth}p{0.2\linewidth}p{0.15\linewidth}
}
    \toprule
    \textbf{Name} & \textbf{Type} &  \textbf{Setting} &  \textbf{Size} & \textbf{Annotations} \\
    \addlinespace[2pt]
    \midrule
    
    Sciortino et al. \cite{sciortino2017estimation} & Video & SSBD dataset + youtube & $1176$ images of $104$ subjects & 2D pose keypoints \\
    \midrule
    
    DREAM \cite{billing2020dream} & Video & Interactions with robot \newline No raw data, only extracted features and annotations & $306$ hours of therapy ($102$) subjects & 3D pose keypoints \\
    \midrule
    
    BabyPose \cite{migliorelli2020babypose} & Video \newline Depth & Preterm Infant movement in NICUs   & $16000$ frames $\cdot$ $16$ depth videos $\cdot$ $16$ patients & 2D pose keypoints \\
    \midrule
    
    SyRIP \cite{huang2021invariant} & Image & Hybrid: real + synthetic \newline YouTube and Google images & Real: $700$ images ($140+$ subjects) \newline Synthetic: $1000$ images & 2D pose keypoints \\
    \midrule
    
    MINI-RGBD \cite{hesse2018computer} & Video \newline Depth & Synthetic: obtained by registering SMIL to real sequences of moving infants. \newline Constrained environment & $12000$ frames $\cdot$ $12$ sequences & 2D and 3D keypoints \\

    \bottomrule
\end{tabular}
\caption{Summary of selected pose estimation children datasets.}
\label{tab: children-datasets}
\end{table*}

\begin{figure*}[th]
    \centering
    \includegraphics[width=\textwidth]{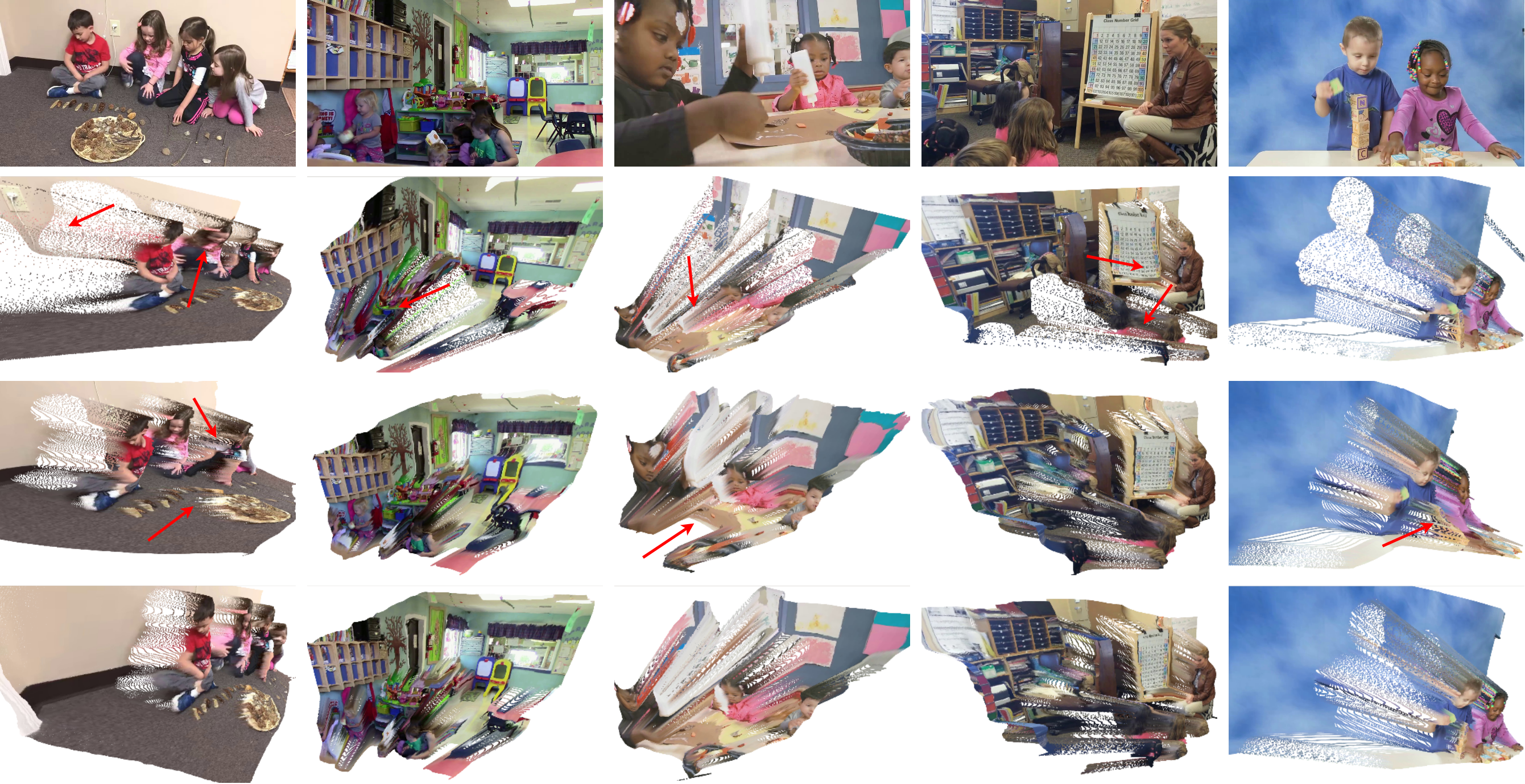}
    \caption{Comparison of point clouds generated using the depth maps from Ranftl et al.~\cite{ranftl2019midas} (row 2), Wei et al.~\cite{Wei2021leres} (row 3) and Patakin et al.~\cite{patakin2022depth} (row 4) on ChildPlay images. The point clouds generated using Patakin et al. appear to best preserve the geometry of the scene.}
    \label{fig:point-cloud-comparison}
\end{figure*}

\mypartitle{Qualitative Results.}
We provide a qualitative comparison of point clouds generated using the depth maps from Ranftl et al.~\cite{ranftl2019midas}, Wei et al~\cite{Wei2021leres} and Patakin et al.~\cite{patakin2022depth} in Figure~\ref{fig:point-cloud-comparison}. We observe that the point clouds generated using the depth maps from Wei et al. and Patakin et al. generally have less distortion of scene elements, and better maintain the depth between objects. The point clouds from Patakin et al. in particular seem to preserve the geometry of the scene best.

\mypartitle{Gaze Vector Stability.}
To quantitatively compare the methods of Wei et al.~\cite{Wei2021leres} and Patakin et al.~\cite{patakin2022depth}, we investigate which algorithm generates more stable gaze vectors. This is crucial as we rely on their generated gaze vectors as ground truth. The test is based on the fact that the gaze vector for a person (camera coordinate system) should be the same irrespective of their distance from the camera. We perform the test as follows:

\begin{itemize}
    \item We take 5 random crops of an image
    \item For each crop, we compute the depth (Wei et al. or Patakin et al.) and focal length
    \item We then reconstruct the point cloud $\pointcloud^{\cam}$ following the protocol defined in Section 4.2, and obtain the gaze vector for each crop as $\gtgaze^{\cam} = \frac{\pointcloud^{\cam}_{gaze} - \pointcloud^{\cam}_{eye}}{||\pointcloud^{\cam}_{gaze} - \pointcloud^{\cam}_{eye}||}$
    \item The stability is given by the standard deviation of the gaze vector across the crops
\end{itemize}

For a more robust estimate, we perform this procedure for the first frame of every clip in the ChildPlay training set, and compute the median standard deviation. The values for the method of Wei et al. are [0.041, 0.032, 0.095] while the values for the method of Patakin et al. are [0.026, 0.019, 0.075]. The median standard deviation for Patakin et al. is lower, especially for the z component, indicating that it generates more stable gaze vectors.

\subsection{Training Details}

\mypartitle{Head Bounding Boxes.}
The provided head box annotations for GazeFollow are not consistent and sometimes include the whole head, and at other times just the face of the person. Hence, we re-extract the head boxes using a pre-trained Yolov5 model~\cite{glenn_jocher_2022_7002879_yolov5} and use these for all our experiments.

\mypartitle{Input Aspect Ratio.}
Previous methods~\cite{chong2020dvisualtargetattention}\cite{gupta2022modular} distort the scene and head images to the model input size. To avoid this, we expand the head bounding box to a square to match the Human-Centric module's input aspect ratio. We also carefully crop and pad scene images to the Scene-Centric module's input aspect ratio during training and validation so that there is no distortion. During the test phase, we do not perform any cropping/padding and instead scale the longer side of the scene image to the Scene-Centric module's input width.

\end{document}